\documentclass[10pt,twocolumn,letterpaper]{article}
\pdfoutput=1
\usepackage{cvpr}
\usepackage{times}
\usepackage{algpseudocode}
\usepackage{algorithm}
\usepackage{epsfig}
\usepackage{graphicx}
\usepackage{amsmath}
\usepackage{amssymb}
\usepackage{amsthm}
\usepackage{caption}
\usepackage{subcaption}
\usepackage[pagebackref=true,breaklinks=true,letterpaper=true,colorlinks,bookmarks=false]{hyperref}

\newtheoremstyle{mydefstyle}{}{}{\itshape}{}{\bfseries}{:}{.5em}{#1 #2 (\thmnote{#3})}
\theoremstyle{mydefstyle}
\newtheorem{mydef}{Definition}
\newtheorem{mytheory}{Theorem}
\newtheorem{mycorr}{Corollary}
\def\real{\mathbb{R}}
\def\posnat{\mathbb{N}^+}
\def\labs{\mathbb{N}}
\newcommand{\argmin}{\operatornamewithlimits{argmin}}
\newcommand{\argmax}{\operatornamewithlimits{argmax}}

\makeatletter
\renewcommand\section{\@startsection {section}{1}{\z@}%
                                   {-1.4ex \@plus -2ex \@minus -.2ex}%
                                   {.5ex \@plus.2ex}%
                                   {\normalfont\Large\bfseries}}
\renewcommand\subsection{\@startsection{subsection}{2}{\z@}%
                                   {-1.2ex \@plus -2ex \@minus -.2ex}%
                                   {.5ex \@plus.2ex}%
	                           {\normalfont\large\bfseries}}
\renewcommand\subsubsection{\@startsection{subsubsection}{3}{\z@}%
                                     {-.5ex\@plus -.2ex \@minus -.2ex}%
                                     {.2ex \@plus .2ex}%
                                     {\normalfont\large\bfseries}}

\def\tightmath{
\abovedisplayskip=4pt plus 2pt minus 1pt 
\abovedisplayshortskip=2pt plus 1pt minus 1pt 
\belowdisplayskip=4pt plus 2pt minus 1pt 
\belowdisplayshortskip=2pt plus 1pt minus 1pt }
\def\crushmath{
\abovedisplayskip=1pt plus 1pt minus 2pt 
\abovedisplayshortskip=1pt plus 1pt minus 2pt 
\belowdisplayskip=1pt plus 1pt minus 2pt 
\belowdisplayshortskip=1pt plus 1pt minus 2pt }
\crushmath

\makeatother

\def\cvprPaperID{557} 

\cvprfinaltrue
\ifcvprfinal\fi
\begin{document}
\tightmath

\title{Towards Open World Recognition}

\author{Abhijit Bendale, Terrance Boult\\
University of Colorado at Colorado Springs\\
{\tt\small \{abendale,tboult\}@vast.uccs.edu}
}

\maketitle

\begin{abstract}
	With the of advent rich classification models and high computational power visual recognition systems have found many operational applications. Recognition in the real world poses multiple challenges that are not apparent in controlled lab environments. The datasets are dynamic and  novel categories must be continuously detected and then added. At prediction time, a trained system has to deal with myriad unseen categories.  Operational systems require minimum down time, even to learn. To handle these operational issues,  we present the problem of \textbf{Open World recognition} and formally define it. We prove that thresholding sums of monotonically decreasing functions of distances in linearly transformed feature space can balance ``open space risk'' and empirical risk. Our theory extends existing algorithms for open world recognition.  We present a protocol for evaluation of open world recognition systems. We present the Nearest Non-Outlier (NNO) algorithm which  evolves model efficiently,  adding object categories incrementally while detecting outliers and  managing open space risk.  
    We perform experiments on the ImageNet dataset with 1.2M+  images to validate the effectiveness of our method on large scale visual recognition tasks.  NNO consistently yields superior results on open world recognition.

\end{abstract}

\section{Introduction} \label{ref:intro}

\begin{figure*}
\centering
\includegraphics[width=1.75\columnwidth]{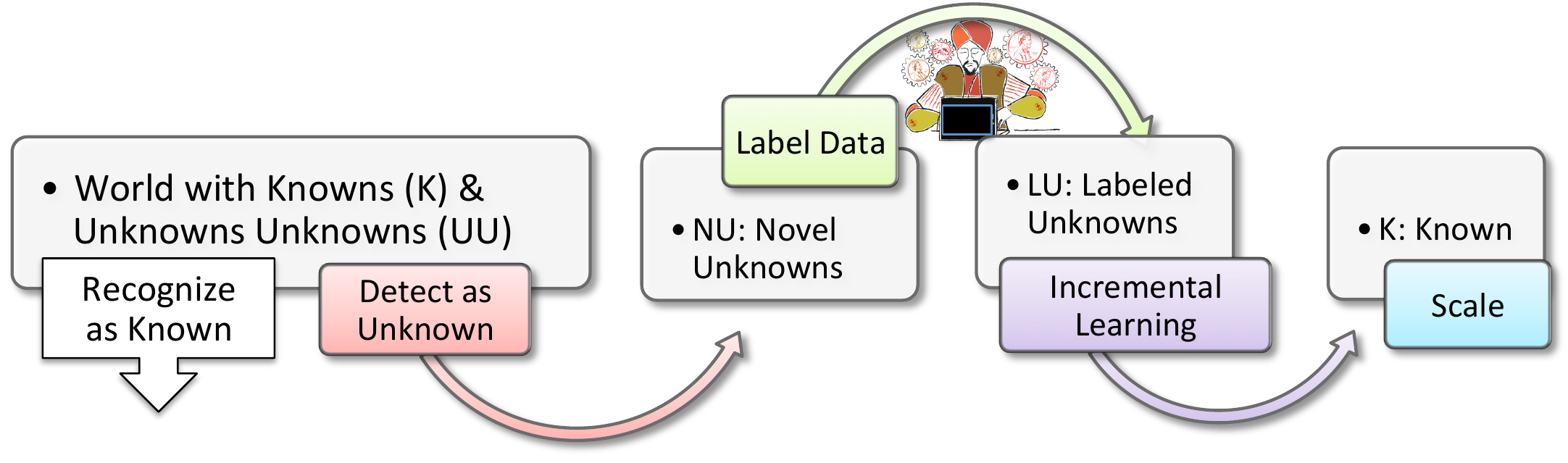}
\caption{ In  open world recognition the system  must be able to recognize objects and associate them with known classes while also being able to label classes as unknown. These ``novel unknowns''  must then be collected and labeled (e.g. by humans).  When there are sufficient labeled unknowns for new class learning, the system must incrementally learn and extend the multi-class classifier, thereby making  each new class ``known'' to the system.   
Open World recognition moves beyond just being robust to unknown classes and toward a scaleable system that is adapting itself and learning in an open world.
}
\label{fig:openworld}
\end{figure*}

 Over the past decade, datasets for building and evaluating visual recognition systems have increased both in size and variation. 
 The size of datasets has increased from a few hundred images to millions of images and the number of categories within the
  datasets has increased from tens of categories to more than a thousand categories. Co-evolution of rich classification models
 along with advances in datasets have resulted in many commercial applications \cite{google, facebook, ebay}.
 A multitude of operational challenges are posed while porting recognition systems from controlled lab environments to real world.
 A recognition system in the ``open world'' has to continuously update with additional object categories, be robust to unseen categories and have minimum downtime.  Despite the obvious dynamic and open nature of the world, a vast majority of recognition systems assume a  \textit{static} and closed world model of the problem where all categories are known a priori. To address these
 operational issues, this paper formalizes and presents steps towards the problem of open world recognition. The key steps  
 of the problem are summarized in  Fig.~\ref{fig:openworld}.  

As noted by \cite{boult-openset}, ``when a recognition system is trained and is operational, there are finite set of known objects in scenes with  myriad unknown objects, combinations and 
 configurations -- labeling something new, novel or unknown should always be a valid outcome''.   One reason for the domination of ``closed world'' assumption of today's vision systems 
 is that matching, learning and classification tools have been formalized as selecting the most likely class from a closed set.
Recent research, \cite{boult-openset, boult-openset-probab, boult-pisvm}, has re-formalized learning for recognition as open set recognition. However, this approach does not 
explicitly require that inputs be  as known or unknown. In contrast, for open world recognition, 
we propose the system to explicitly label novel inputs as unknown and then incrementally incorporate them into the classifier.
Furthermore, open set recognition as formulated by \cite{boult-openset} is designed for
traditional one-vs-all batch learning scenario. Thus, it is open set but not incremental and does not scale gracefully
with the number of categories.

While there is a significant body of  work on incremental learning algorithms  that handle new instances of known classes \cite{poggio-nips, crammer-passive, yeh-cvpr08}, 
open world requires two more general and difficult steps: continuously detecting novel classes and when novel inputs are found updating the system to include these new classes in its multi-class open set recognition algoritim.  
Novelty detection and outier detection are complex issues in their own right with long histories \cite{markou2003novelty,hodge2004survey}  and are still  active vision research topics \cite{bodesheim2013kernel,lu2012image}.   After detecting a novel class, the requirement to add new classes leaves the system designer with the choice of re-training the entire system. When the number of categories are small, such a solution may be feasible, but unfortunately, it does not scale. Recent studies on ImageNet dataset using SVMs or CNN require days to train their system  \cite{perronnin_towards_2012, krizhevsky2012imagenet}, e.g. 5-6 CPU/GPU days in case of CNN for 1000 category image classification task. 
Distance based classifiers like  Nearest Class Mean (NCM)  \cite{kapoor-cvpr, mensink, ristin-ncmf} offer a natural choice for building scalable system that can learn new classess incrementally. In NCM-like classifiers, incorporating new images or classes in implies adjusting the existing means or updating the set of class means. 
However, NCM classifier in its current formulation is not suited for open set recognition because it uses close-set assumptions for probability normalization. Handling unknowns in open world
recognition requires  gradual decrease in the value of probability (of class membership) as the test point moves away from known
data into open space. The Softmax based probability assignment used in NCM does not account for open space.

The {\em first contribution} of this paper is a formal definition of the problem of open world recognition, which extends the existing 
definition of open set recognition which was defined for a static notion of set. In order solve open world recognition, the system needs 
be robust to unknown classes, but also be able to move through the stages and knowledge progression summarized in Fig.~\ref{fig:openworld}.   {\em Second contribution} of the work is a recognition system that can continuously learn new object categories in an open world model.  In particular, we  show how to extend Nearest Class Mean type algorithms (NCM) \cite{mensink}, \cite{ristin-ncmf},  to a Nearest Non-Outlier (NNO) algorithm that can balance open space risk and accuracy.  

To support this extension, our {\em third contribution} is  showing that thresholding  sums of monotonically decreasing functions of distances of linearly
transformed feature space can have arbitrarily small ``open space risk''.
Finally,  we present a protocol for evaluation for open world recognition, and use  this protocol to show our NNO algorithm perform significantly better on open world recognition evaluation using Image-Net \cite{berg-imagenet}.

\section{Related Work} \label{related-work}

	Our work addresses an issue that is related to and has received attention from various communities such as incremental learning, scalable and open set learning. 
	
	\textbf{Incremental Learning:} As SVMs rose to prominence in many object recognition \cite{svm-knn, lin-fastsvm-cvpr11}, many incremental extensions to SVMs were proposed. Cauwenberghs \etal \cite{poggio-nips} proposed an incremental binary SVMs
	by means of saving and updating KKT conditions.  Yeh \etal \cite{yeh-cvpr08}  extended the approach to object recognition and demonstrated multi-class incremental learning. Pronobis \cite{caputo-memcontrol} proposed memory-controlled online incremental SVM for visual place recognition.
	Although incremental SVMs might seem like a natural for large scale incremental learning for object recognition, they suffer from multiple drawbacks. The update process is extremely expensive (quadratic in the number of
	training examples learned \cite{laskov}) 
	and depends heavily on the number of support vectors stored for performing updates \cite{laskov}.
	To overcome the update expense, \cite{crammer-passive} and \cite{pegasos} proposed classifiers with fast and inexpensive update process along with their multi-class extensions.
	However, the multi-class incremental learning methods and other incremental classifiers, \cite{crammer-passive, pegasos, multi-pegasos, li-cvpr07}, are incremental in terms of additional training samples. 
		
\begin{figure}[t]
\centering
\includegraphics[width=1.1\columnwidth]{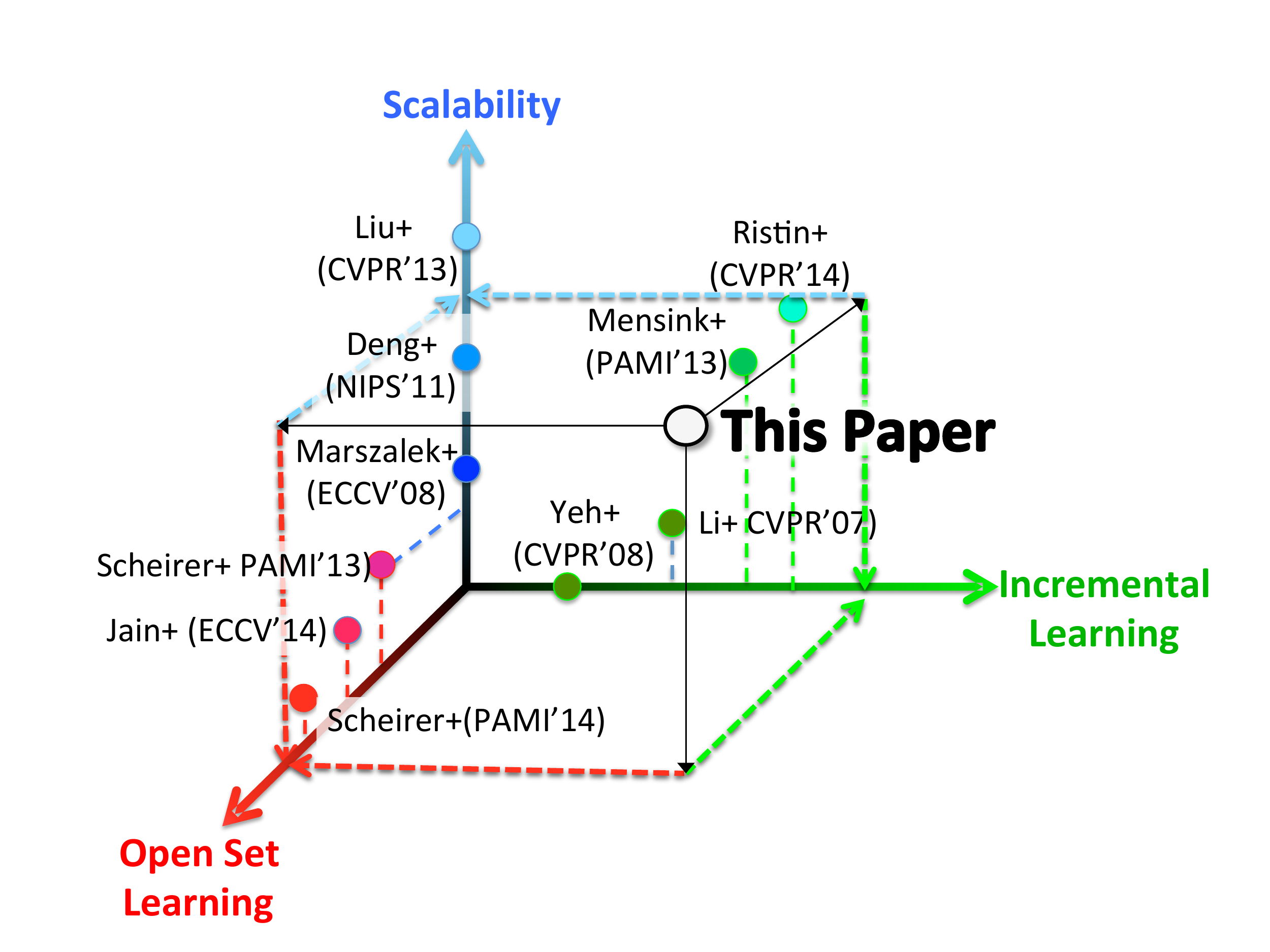}
\caption{ Putting the current work in context by  depicting locations of prior work with respect to three axes of the major issues for open world recognition: open set learning, incremental learning and scalability. In this work, we present a system that is scalable, can handle open set recognition and can learn new categories
incrementally without having to retrain the system every time a new category arrives.
The works depicted include Ristin \etal \cite{ristin-ncmf}, Mensink \etal \cite{mensink}, Scheirer \etal. \cite{boult-openset}, \cite{boult-openset-probab},
Jain \etal. \cite{boult-pisvm}, Yeh \etal, \cite{yeh-cvpr08}, Marszalek \etal. \cite{schmid-2008}, Liu \etal \cite{liu-2013}, Deng \etal \cite{jia-deng}, and Li \etal \cite{li-cvpr07}. This papers advances the state of the art in open set learning, in incremental learning while providing reasonable scalability. }
\label{fig:teaser}
\end{figure}

	\textbf{Scalable Learning:} Researchers like \cite{schmid-2008, liu-2013, jia-deng} have proposed label tree based classification methods to address
	scalability (\# of object categories) in large scale visual recognition challenges  \cite{pascal, berg-imagenet}. Recent advances in deep learning community 
	\cite{hinton-nips12}, \cite{simonyan-arxiv} has resulted in state of the art performance on these challenges. Such methods are extremely useful when the goal is obtain 
	maximum classification/recognition performance. These systems assume a priori availability of entire training data (images and categories). 	
	However,  adapting such methods to a dynamic learning scenario becomes extremely challenging. Adding object categories requires retraining the entire system, 
	which could be infeasible for many applications. Thus, these methods are scalable but not incremental (Fig ~\ref{fig:teaser})

	\textbf{Open Set Learning:} Open set recognition assumes that there is  incomplete knowledge of the world is present at training time, and 
	that unknown classes can be submitted to an algorithm during testing \cite{transduction-pami, boult-openset}. Scheirer \etal \cite{boult-openset} formulated the problem of open set
	recognition for static one-vs-all learning scenario by balancing open space risk while minimizing empirical error.  Scheirer \etal \cite{boult-openset-probab, boult-pisvm} extended 
	the work to multi-class settings by introducing compact abating probability model. Their work offers insights into building robust methods to handle unseen categories.
	However, class specific Weibull based calibration of SVM decision scores does not scale. Fragoso \etal \cite{evsac} proposed a scalable Weibull based calibration for hypothesis
	generation for modeling matching scores, but do not address it in the context of general recognition problem.

The final aspect of related work is nearest class mean (NCM) classifiers.  NCM classification, in which samples undergo a Manhanalobis
transform and then are are associated with a  class/cluster mean,  is a  classic pattern recognition approach
\cite{fukunaga1990introduction}.  NCM classifiers have a long history of use in vision
systems, \cite{crisman1989color} and have multiple extensions, adapations and applications \cite{datta1997symbolic,skurichina2002bagging,yang2006music,kuncheva2008case,loog2010constrained}.  Recently the technique has been adapted for
use in larger scale vision problems \cite{veenman2005weighted,veenman2005nearest,mensink,ristin-ncmf}, with
the most recent and most accurate approaches combining NCM with metric learning \cite{mensink} and with random forests\cite{ristin-ncmf}.  

Since we extend NCM classification, we briefly review the formulation including a probabilistic interpretation. Consider an image
represented by a $d$-dimensional feature vector $x \in \mathbb{R}^d$. Consider
$\mathcal{K}$ object categories with their corresponding centroids $\mu_k$,
where $k \in \mathcal{K}$. Let $\mathcal{I}_k$ be images for each object
category.  The centroid is given by $\mu_k = \frac{1}{|\mathcal{I}_k|} \sum_{i \in \mathcal{I}_k}x_i$. 
NCM classification of a given image instance $I$ with a feature vector $x$ is formulated as searching for the closest centroid in feature space as
$c^* = \underset{k \in \mathcal{K}} {\mathrm{argmin}} \; \; \mathbf{d} (x , \mu_k)$.
Here $\mathbf{d}(.)$ represents a distance operator usually in  Euclidean space. Mensink \etal  \cite{mensink} replace Euclidean distance with a low-rank Mahalanobis distance optimized on training data. The Mahalanobis distance
is induced by a weight matrix $W \in \mathbb{R}^{d \times D}$, where D is the dimensionality of the lower dimensional space. Class conditional probabilities $p(c|x)$ using an NCM classifier are obtained using a probabilistic
model based on multi-class logistic regression as follows:
\begin{equation} 
p(c|x) = \frac{exp(- \frac{1}{2} \mathbf{d}_W (x, \mu_k))}{ \sum_{k' =1}^{\mathcal{K}} exp(- \frac{1}{2} \mathbf{d}_W (x, \mu_{k'})) }
\label{eq:ncmprob}
\end{equation}

In the above formulation, class probabilities $p(c)$ are set to be uniform over all classes. During Metric learning optimization,  Mensink \etal \cite{mensink} considered non-uniform probabilities given by:
\begin{equation}
p'(c|x) = \frac{1}{Z}exp(x^TW^TW \mu_c + s_c)
\label{eq:ncm-dotprob}
\end{equation}
where Z denotes the normalizer and $s_c$ is a per class bias.

\section{Open World Recognition} 
We first establish preliminaries related to open world recognition, following which we formally define the problem.
Let classes be labeled by positive integers $\posnat$ and let ${\cal K}_t \subset \posnat$ be the set of labels of known
classes at time $t$.  Let the zero label ($0$) be reserved for (temporarily) labeling
data as unknown. Thus $\labs$ includes unknown and known labels. 

Let our features be $x\in \real^d$.  Let $f\in {\cal H}$ be a measurable
recognition function, i.e. $f_y(x)>0$ implies recognition of the
class $y$ of interest and $f_y(x)\le 0$ when $y$ is not recognized, where 
${\cal H}:\real^d\mapsto\real$ is a suitably smooth space of recognition functions.

The objective function of open set recognition, including multi-class
formulations, must balance open space risk against empirical error.  As a
preliminary we adapt the definition of open space and open space risk used
in~\cite{boult-openset}.  Let open space, the space sufficiently far from any
known positive training sample $x_i\in {\cal K}, i = 1 \ldots N$, be defined as:
\begin{equation}
{\cal O} = S_o - \bigcup_{i \in N}{B_r(x_i)}
\label{eq:openspace}
\end{equation}
where $B_r(x_i)$ is a closed ball of radius $r$ centered around any training sample $x_i$.
Let $S_o$ be a ball of radius $r_o$ that includes all 
known positive training examples $x \in {\cal K}$ as well as the open space
$\cal O$. Then  probabilistic {\em Open Space Risk} $R_{\cal O}(f)$ for a class
$y$ can  be defined as
\crushmath
\begin{equation}
R_{{\cal O}}(f) =  \frac{\int_{{\cal O}} f_y(x) dx}{\int_{S_o} f_y(x) dx} 
\label{eq:openrisk}
\end{equation}
That is, the open space risk is considered to be the relative measure of positively
labeled open space compared to the overall measure of positively labeled space.

Given an empirical risk function $R_{\cal E}$, e.g. hinge loss, the objective of {\em open set recognition} is to find a measurable recognition function that manages (minimizes) the {\bf Open Set Risk}:
\begin{equation}
\label{eq:minbayeserror}
\argmin_{f \in {\cal H}} \left\{R_{\cal O}(f)  + \lambda_r R_{\cal E}(f)\right\}
\end{equation}
where $\lambda_r$ is a regularization constant.   
\tightmath

With the background in place, we formalize the problem of  open world recognition. 

\begin{mydef}[Open World Recognition]
\setlength\abovedisplayskip{0pt}
\setlength\belowdisplayskip{2pt}
A solution to open world recognition is a tuple $[F, \varphi, \nu, L, I]$ with:
\label{def:openworld}
\begin{enumerate}
\item 
 A  {\bf  multi-class open set recognition function $F(x) :\real^d \mapsto \labs$} using a vector function $\varphi(x)$ of $i$ per-class measurable recognition functions $f_i(x)$,   also using  a {\bf novelty detector  $\nu(\varphi):\real^i \mapsto [0,1]$}.   We require the per class recognition functions $f_i(x) \in {\cal H} :\real^d \mapsto \real$ for $i\in {\cal K}_t$  to  be open set recognition functions that manage open space risk as Eq.\ref{eq:openrisk}.   The novelty detector  $\nu(\varphi):\real^i \mapsto [0,1]$  determines if results from vector of  recognition functions is from an unknown ($0$) class. 
\item A labeling process $L(x):\real^d \mapsto \posnat$ applied to novel unknown data  $U_t$ from time $t$, yielding labeled data $D_t = \{ (y_j,x_j)\}$ where $y_j=L(x_j) \forall x_j \in U_t$. Assume the labeling  finds $m$ new classes, then  the set of known classes becomes  ${\cal K}_{t+1}={\cal K}_t\cup \{ i+1, \ldots i+m \} $.
\item An incremental learning function $I_t({\varphi};D_t): {\cal H}^i\mapsto{\cal H}^{i+m}$ to scaleably learn and add new measurable functions $f_{i+1}(x)\ldots f_{i+m}(x)$, each of which  manages open space risk, to the vector $\varphi$ of measurable recognition functions.
\end{enumerate} 
\end{mydef}

Ideally all of these steps should be automated, but herein we presume supervised
learning with labels obtained by human labelling.

If we presume that each $f_k(x)$ reports a likelihood of being in class $k$ and
that we presume $f_k(x)$ normalized across the respective classes and 
Let $\varphi = [ f_1(x), \ldots, f_k(x) ]$.  For this paper we let
the multi-class open set recognition function be given as
\begin{align}
y^* &= \argmax_{y\in {\cal K}, f_y(x) \in \varphi(x)} f_y(x),  \\
F(x) &= \begin{cases} 
  0 & \hbox{if } \nu(\varphi(x)) = 0 \\
 y^*  & otherwise 
  \end{cases}
\label{eq:multiclass}
\end{align}
With these definitions a simple approach for the novelty detection is to set a minimum threshold $\tau$ for acceptance, e.g. 
letting  $\nu(\varphi(x)) = f_{y^*}(x) > \tau$.  In the following section we will prove this simple approach can manage open space risk and hence provide for item 1 in the open world recognition definition.

\section{Opening existing algorithms}

The series of papers \cite{boult-openset,boult-openset-probab,boult-pisvm}
formalized the open set recognition problem and proposed 3 different algorithms
for managing open set risk.  It is natural to consider these algorithms
for open world recognition.  Unfortunately, these algorithms use EVT-based calibration of 1-vs-rest RBF SVMs and hence are not
well suited for incremental updates or scalability required for open world recognition.
In this paper we pursue an alternative approach better suited to
open world using non-negative combinations of abating distance.  Using this
approach Sec ~\ref{sec:NNO}  shows that NCM can be inexpensively
extended to open world recognition, which is termed as Nearest Non-Outlier (NNO) algorithm.

The authors of \cite{boult-openset-probab}  show that if a recognition function is decreasing away from the training data, a property they call abating, then thresholding the abating function limits the labeled region and hence can manage/limit open space risk.  The Compact Abating Probability (CAP) model presented in that paper is a sufficient model, but it is not necessary.   In particular we build on the concept of a CAP model but generalize the model showing that any non-negative combination of abating functions, e.g., a convex combination of decreasing functions of distance, can be thresholded to have zero open space risk.  We further show that we can work in linearly transformed spaces, including projection onto subspaces, and still manage open space risk and that NCM type algorithms manage open space risk.

\begin{mytheory}[\bf Open space risk for model combinations]
Let $M_{\tau,y}(x)$ be a recognition function that thresholds a non-negative
weighted sum of $\eta$ CAP models ($M_{\tau,y}(x) = \sum_{j=1}^{\eta} c_j
M_{j,\tau_j,y}(x) $ ) over a known training set for class $y$, where $1 \ge
c_j\ge 0$ and $M_{j,\tau,y}(x)$ is a CAP model. Then for $\delta \ge 0\ \exists
\tau^* $ s.t. $R_{{\cal O}}(M_{\tau^*,y}) \le \delta$, i.e. one can threshold
the probabilities $M_{\tau,y}(x)$ to limit open space risk to any desired level.
\label{theory:minrisk}
\end{mytheory}

Proof:  It is sufficient to show the condition holds for $\delta=0$, since similar to Corollary  1 of \cite{boult-openset-probab}, larger values of $\delta$ may simply allow larger labeled regions with larger open space risk. Considering each model $M_{j,\tau_j,y}(x) j=1..\eta$ separately, we can apply Theorem 1 of \cite{boult-openset-probab} to each $M_{j,\tau_j,y}(x)$  yielding a $\tau_j$ such that the function $M_{j,\tau_j,y}(x)>0$ defines a labeled region $l_j(\tau_j)\subset X$ with zero open space risk. Letting $\tau^* = \min_j \tau_j$ it follows that $M_{\tau^*,y}(x)>0$ is contained within $\cup_j l_j(\tau^*)$, which as a finite union of compact regions with zero risk, is itself a compact labeled region with zero open space risk. {\em Q.E.D}

The theorem/proof trivially holds for a max over classes. The proof can be generalized to combinations via product.
The proof can also be generalized to combinations of monotonic transformed recognition functions, with appropriate choice of thresholds, but for
 this paper we need only a max or sum of models.  However, we also need to work in transformed spaces especially in lower-dimensional projected spaces.

\begin{mytheory}[\bf Open Space Risk for Transformed Spaces]
Given a linear transform $T:\real^n\to\real^m$ let  $x' = T(x), \forall x \in X$, yields $X'$  a linearly transformed  space of features derived from feature space $X\subset\real^n $.  Let ${\cal  O'} = \cup_{x \in {\cal  O}} T(x) $ be the transformation of  points in open space ${\cal  O}$.   Let $M'_{\tau,y}(x') $ be a probabilistic CAP recognition function over $x'\in X'$ and let $M_{\tau,y}(x) = M'_{\tau,y}(Tx)$ be a recognition function over $x\in X$.    Then  $\exists \epsilon : R_{{\cal  O'}}(M'_{\tau',y}) \le \delta \implies R_{{\cal  O}}(M_{\tau,y}) < \epsilon \delta$, i.e. managing  open set risk in $X'$ will also manage it in the original feature space $X$. 
\label{theory:transform}
\end{mytheory}

Proof:  If $T$ is dimensionalty preserving, then the theorem follows from the linearity of integrals in the definition of risk.  Thus we presume  $T$ is projecting away ${n-m}$ dimensions.  Since the open space risk in the projected space is $\delta$ we have $\lambda_{m}(M'_{\tau',y}\cap{\cal O'})= c \delta$ where $\lambda_{m}$ is the Lebesgue measure in $\real^m$ and $c<\infty$.   Since ${\cal O}\subset S_o$, i.e. ${\cal O}$ is contained within a ball of radius $r_o$,  it follows from the properties of Lebesgue measure that $ \lambda_n (M_{\tau,y}\cap{\cal O}) \le \lambda_m \left(M'_{\tau',y} \cap ({\cal O'} \times [-r_o,r_o]^{n-m})\right) = c * \delta * (2r_o)^{n-m}=0$ and hence the open space risk in $\real^m$ is bounded. {\em Q.E.D}.

It is desirable for open world problems that we consider the error in the original space.  We note that $\epsilon$ varies with dimension and the above bounds are generally not tight.  While the theorem gives a clean bound for zero open space risk, for a solution with non-zero $\delta$ risk in the lower dimensional space, when considered in the original space,  the solution may have open space risk that increases exponentially with the number of missing dimensions.

We note that these theorems are not a license to claim that algorithms, with rejection, manage open space risk. While many algorithms can be adapted to compute a probability estimate of per class inclusion and can threshold those probabilities to reject, not all such algorithms/rejections manage open space risk.  Thresholding Eq ~\ref{eq:ncm-dotprob}, which \cite{mensink} minimizes in place of ~\ref{eq:ncmprob}, will not work because the function does not always decay away from  known data.  Similarly, rejecting decision close to the plane in a linear SVM does not manage open space risk, nor does the thresholding layers in a convolution neural network \cite{overfeat}.

On the positive side,  these theorems show that one can adapt algorithms that linearly transforms feature space and use a {\em probability/score mapping that combines positive scores that decrease with distance from a finite set of known samples}. In the following section, we demonstrate how to generalize an existing algorithm while managing open space risk. Open world performance, however, greatly depends on the underlying algorithm and the rejection threshold.  While theorems 1 and 2 say there exists a threshold with zero open space risk, at that threshold there may be minimal or no generalization ability.

\subsection{Nearest Non-Outlier (NNO)} \label{sec:NNO}

As discussed previously (sec ~\ref{ref:intro}), one of the significant contributions of this paper is  combining  theorems ~\ref{theory:minrisk} and ~\ref{theory:transform} to provide an example of  open space risk management  and move toward a solution to open world recognition.  
Before moving on to defining open world NCM, we want to add a word of caution about ``probability normalization'' that  presumes all classes are known.
e.g. softmax type normalization used in eqn ~\ref{eq:ncmprob}.
Such normalization is problematic for 
open world recognition, where there are unknown classes, In particular, {\bf  in open world recognition the Law of Total Probability and Bayes'
  Law cannot be directly applied} and hence cannot be used to normalize
scores.  Furthermore, as one adds new classes, the normalization factors and hence probabilities,
keep changing and thereby  limiting  interpretation of the probability.
For an NCM type algorithm, normalization with the softmax makes thresholding
very difficult since for points far from the class means the nearest mean will have a probability near 1.  Since it does not decay, it does not follow Theorem ~\ref{theory:minrisk}. 

To adapt NCM for open world recognition we introduce Nearest Non-Outlier (NNO) which uses a measurable recognition function consistent with Theorems ~\ref{theory:minrisk} and ~\ref{theory:transform}. 
Let NNO represent its internal model as a vector of means ${\cal M}= [ \mu_{1}, \ldots \mu_{k}]$.  Let $W \in \mathbb{R}^{d \times m}$ be the linear transformation dimensional reduction weight matrix
learned by the process described in  \cite{mensink}.   Then given $\tau$, let  
\begin{equation}
\hat{f}_i(x) =  \frac{\Gamma(\frac{m}{2}+1)}{\pi^{\frac{m}{2}} \tau^m} (1- \frac{1}{\tau}\|  W^\top x - W^\top \mu_{i}\|)
\label{eq:nnodist}
\end{equation}
be our measurable recognition function with $\hat{f}_i(x)>0$ giving the probability of being in class in class $i$, where $\Gamma$ standard gamma function which occurs in the volume of a m-dimensional ball.  Let $\hat{\varphi} = [ \hat{f}_1(x), \ldots, \hat{f}_k(x) ]$ with $y^*$ and $F(x)$  given by Eq.~\ref{eq:multiclass}. Let with  $\hat{\nu}(\hat{\varphi}(x)) = \hat{f}_{y^*}(x) > 0$.  

That is, NNO rejects $x$ as an outlier for class $i$ when $\hat{f}_i(x)=0$, and labels input $x$ as unknown/novel when all classes reject the input.
Finally, after collecting novel inputs,  let $D_t$ the human labeled data for a new class $k+1$  and let our incremental class learning $I_t({\hat{\varphi}};D_t)$  compute  $\mu_{k+1} =mean(D_t)$ and append $\mu_{k+1}$ to ${\cal M}$.

\begin{mycorr}[NNO solves open world recognition]
The NNO algorithm with human labeling $L(x)$ of unknown inputs is a tuple
$[F(x),\hat{\varphi},\hat{\nu}(\hat{\varphi}(x)),L, I_t({\hat{\varphi}};D_t)]$,  consistent with
Definition~\ref{def:openworld}, hence NNO is a open world recognition algorithm. 
\end{mycorr}
By construction theorems 1 and 2 apply to the measurable recognition functions $F(x)$ from
Eq.~\ref{eq:multiclass} when using a vector of per classess functions given
eq.~\ref{eq:nnodist}.  By inspection the NNO definitions
of $\hat{\nu}(\hat{\varphi}(x))$ and $I_t({\hat{\varphi}};D_t)$ are consistent with Definition~\ref{def:openworld} and  are scaleable.  {\em Q.E.D}.

\section{Experiments}

In this section we present our protocol for open world experimental evaluation of NNO, and a comparison to NCM based classifiers.

\textbf{Dataset and Features:} 
Our evaluation is  based on the  ImageNet Large Scale Visual Recognition Competition 2010 dataset. ImageNet 2010 dataset is a large scale dataset with images from 1K visual categories.
The dataset contains 1.2M images for training (with around 660 to 3047 images per class), 50K images for validation and 150K images for testing. Large number of visual categories allow us to effectively gauge performance of incremental and open world learning scenarios.   In order to effectively conduct  experiments using open set protocol, we need access to ground truth. ILSVRC'10 is the only ImageNet dataset will full ground truth, which is why we selected that dataset over later releases of ILSVRC (e.g. 2011-2014).

 We used densely sampled SIFT features clustered into 1K visual words as given by Berg \etal \cite{berg-imagenet}.  Though more advanced features are available \cite{perronnin_towards_2012,krizhevsky2012imagenet, simonyan-nips2013}, extensive evaluation
across features is beyond the scope of this work \footnote{In the supplemental material we present some experiments on additional features on ILSVRC'13 data so show the advantages of NNO are not feature dependent}. Each feature is whitened by its
mean and standard deviation to avoid numerical instabilities. We report performance in terms of average classification accuracy obtained using top-1 accuracy as per the protocol
provided for the ILSVRC'10 challenge. As our work involves initially training a system with small set of visual categories and incrementally adding
additional ategories, we shun top-5 accuracy. 

\textbf{Algorithms:} We use code provided by Mensink \etal \cite{mensink} as the baseline.  This algorithm has near state of the art results and while recent extension with random forests\cite{ristin-ncmf} improved accuracy slightly, \cite{ristin-ncmf} does not provide baseline code.  Since we  are primarily focused on open world aspects, the NCM baseline using the original authors code provide a sufficient baseline. The baseline NCM algorithm is evaluated using closed set (CS-NCM) and open set (OS-NCM) in incremental learning phase. 
We also report performance on our Nearest Non-OUTLIER (NNO)  extension of NCM classifier in both close-set testing (CS-NNO) and Open set testing (OS-NNO).

\subsection{Open World Evaluation Protocol} 

Closed set evaluation is when a system is tested with all objects known during testing, i.e. training and testing use the same classes but different instances.    In open set evaluation, the system is tested with examples from both known and unknown categories, where unknown categories are categories not used during training.  
Open set recognition evaluation protocol proposed by  by Scheirer \etal \cite{boult-openset} does not handle the open world scenario in which object categories are being added to the system continuously.  Ristin \etal \cite{ristin-ncmf} presented an incremental closed set learning scenario where novel object categories are added continuously. We combined ideas from both of these approaches and propose a protocol that is suited for open world recognition in which categories are being added to the system  continuously while the system is also tested with unknown categories.

\textbf{Training Phase:} The training of the NCM classifier is divided into two phases: an initial metric learning/training phase and a
growth/incremental learning phase.  In the metric learning phase, a fixed set of object categories are provided to the system. The system performs parameter optimization including metric learning on these categories. 
Once the metric learning phase is completed, the incremental learning phase uses the fixed metrics and parameters. During the incremental learning phase, object categories
are added to the system one-by-one. While for scaleability one might measure time,  both NCM and NNO add new categories in the same way and it is  extremely fast, since it only consists of computing the means, so we don't report/measure timing here.  

Nearest Non Outlier (NNO) is our extension of NCM classifier based on
the CAP model requires estimation of $\tau$ for eq.~\ref{eq:nnodist}. This is done in the parameter estimation phase using the metric also learned in that phase.  The validation data for training phase is divided in two sets: known categories
and unknown categories. A $\tau$ for NNO is estimated over the
training known and unknown categories by optimizing for F1-measure. This process
is repeated over multiple folds and the average $\tau$ is obtained. During
evaluation process, the average $\tau$ is used and thresholding at zero  determines if the incoming image 
belongs to an unknown category.

\textbf{Testing Phase:}  To ensure proper open world evaluation, we split the ImageNet test data into 
two sets of 500 categories each: the known set  and the unknown set. 
 At every stage, the system is evaluated with a subset of the  known set and the unknown set to obtain closed set and open set performance.
 This process is repeated as we continue to add categories to the system. The whole process is repeat
ed across multiple dataset splits to ensure fair comparisons and estimate error.
 While \cite{boult-openset} suggest a particular openness measure, it does not address the incremental learning paradigm.
 We fixed the number of unknown categories and report performance as series of known categories are incrementally added.  
 We present separate plots for different number of unknown categories. 

Multi-class classification error \cite{crammer-multiclass} for a system $F_\mathcal{K}(.)$ trained with test samples $\{(x_i, y_i)\}_{i=1}^N, y_i \in \mathcal{K}$ is given as $\epsilon_\mathcal{K} = \frac{1}{N} \sum_{i=1}^N  [\![ F_\mathcal{K}(x_i) \neq y_i ]\!]  $
For open world testing the evaluation must keep track of the errors which occur
due to standard multi-class classification over known categories as well as
errors between known and unknown categories.  Consider evaluation of $N$ samples
from $\mathcal{K}$ known categories and $N'$ samples from $\; \mathcal{U}$
unknown categories leading to $(N+N')$ test samples and $ \mathcal{K} \cup
\mathcal{U} \in X$.  Thus, open world error $\epsilon_{OW}$ for a system
$F_\mathcal{K}(.)$ trained over $\mathcal{K}$ categories is given as:
\crushmath
\begin{equation} \label{eqn:owerror}
\epsilon_{OW} = \epsilon_\mathcal{K} +  \frac{1}{N'} \sum_{j=N+1}^{N'}  [\![ F_\mathcal{K}(x_j) \neq unknown ]\!]  
\end{equation}

\subsection{Experimental Results}

\begin{figure*}
        \centering
        \begin{subfigure}[b]{0.3\textwidth}
                \includegraphics[width=\textwidth]{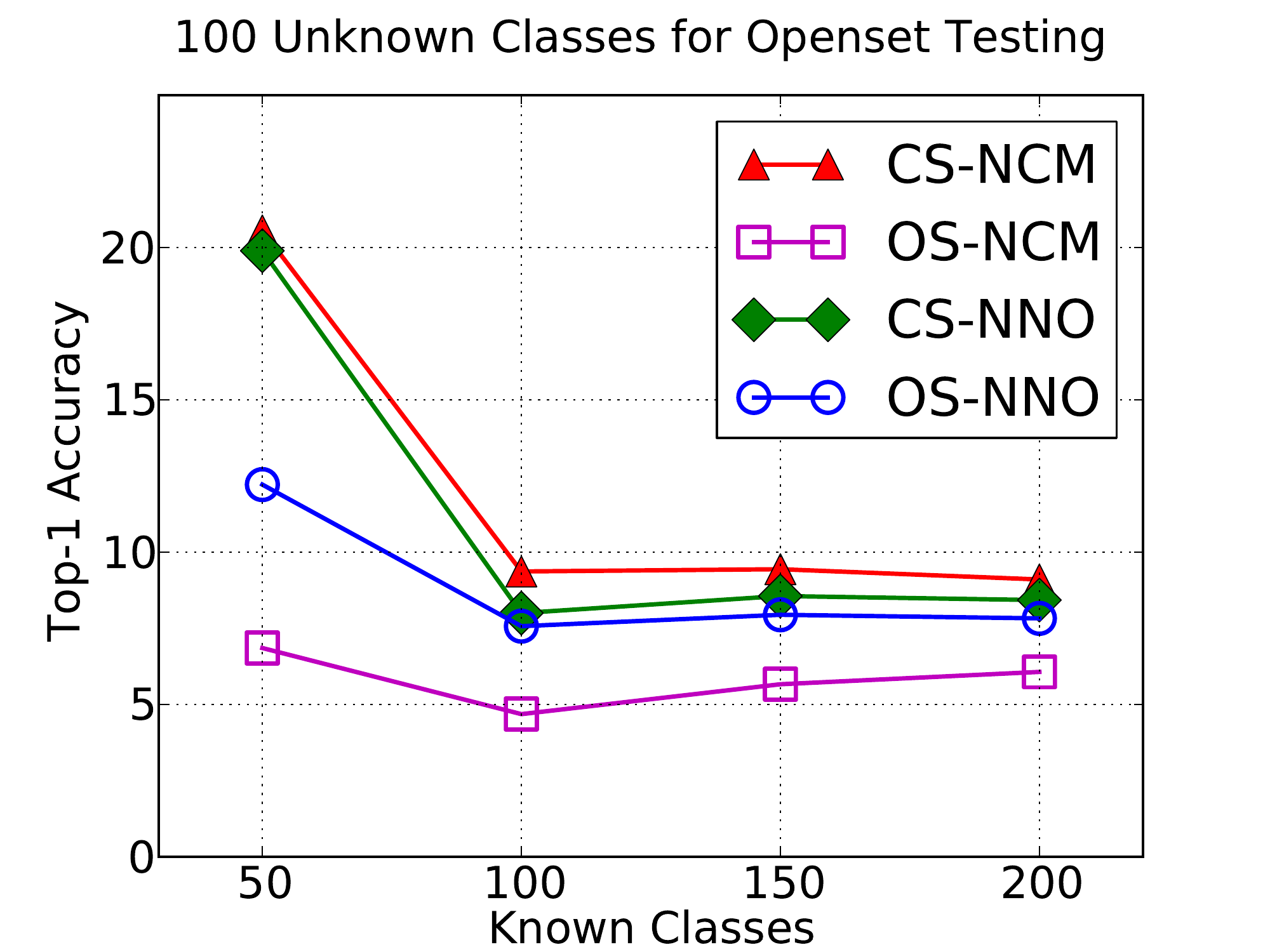}
                \caption{}
                \label{fig:50a}
        \end{subfigure}%
        ~ 
        \begin{subfigure}[b]{0.3\textwidth}
                \includegraphics[width=\textwidth]{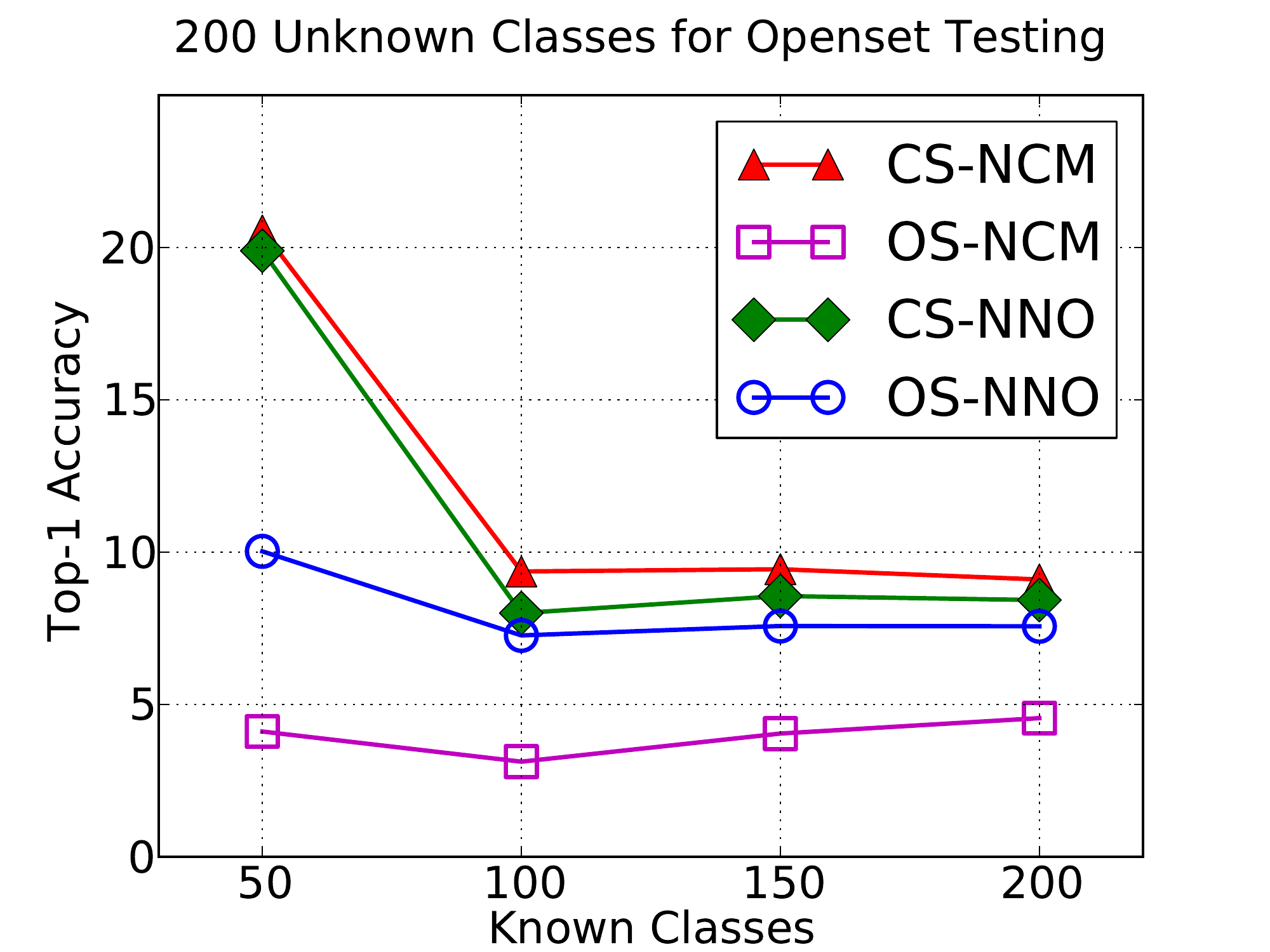}
                \caption{}
                \label{fig:50b}
        \end{subfigure}
        ~ 
        \begin{subfigure}[b]{0.3\textwidth}
                \includegraphics[width=\textwidth]{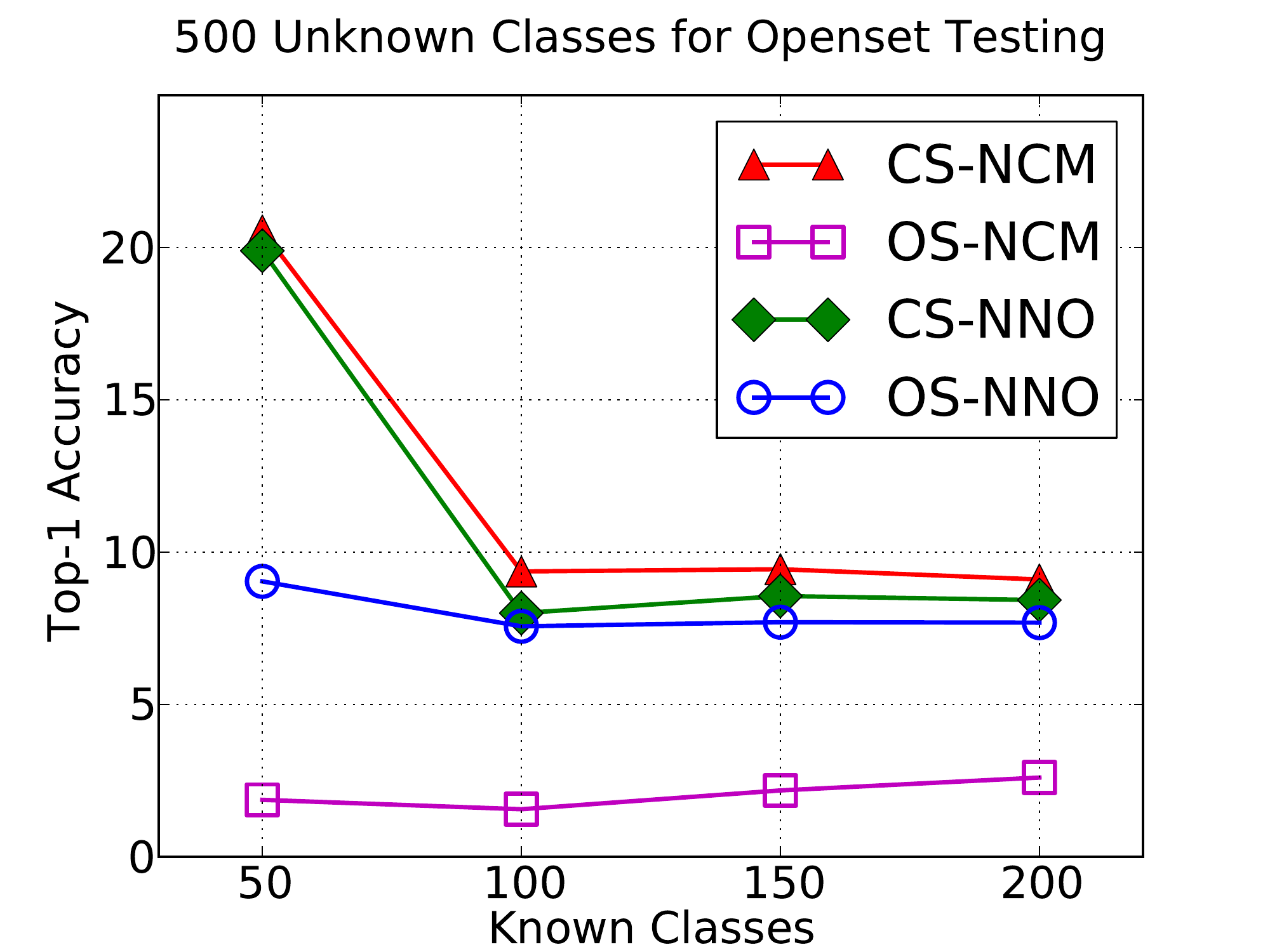}
                \caption{}
                \label{fig:50c}
        \end{subfigure}
        \caption{Open World learning on ILSVRC'10 challenge with 50 initial categories. Top-1 accuracy is plotted as a function of known classes in the system. Note that in all figures CS-NCM is pure close set testing as we vary the number of incrementally learned classes.   The number of unknown categories used for open set testing increases from 100 (Fig.~\ref{fig:50a}) to 200 (Fig.~\ref{fig:50b}) to  500 (Fig.~\ref{fig:50c}). There is significant performance drop
        between closed set testing of NCM (CS-NCM) and open set testing of NCM (OS-NCM). The performance drop increases 
        as the number of unknown categories used for testing increases. Our Nearest Non-Outlier (NNO) approachof handling unknown categories         based on extending NCM with Compact Abating Probabilities, similar results on close set (CS-NNO) and yields significantly better results in open set testing (OS-NNO).  Interestingly, the gap between OS-NNO in open set and  close testing decreases with increaseing classes -- suggesting a correlation between doing well on large scale recongition and robustness to unseen classes. }
        \label{fig:openworld50}
\end{figure*}

\begin{figure*}
        \centering
        \begin{subfigure}[b]{0.3\textwidth}
                \includegraphics[width=\textwidth]{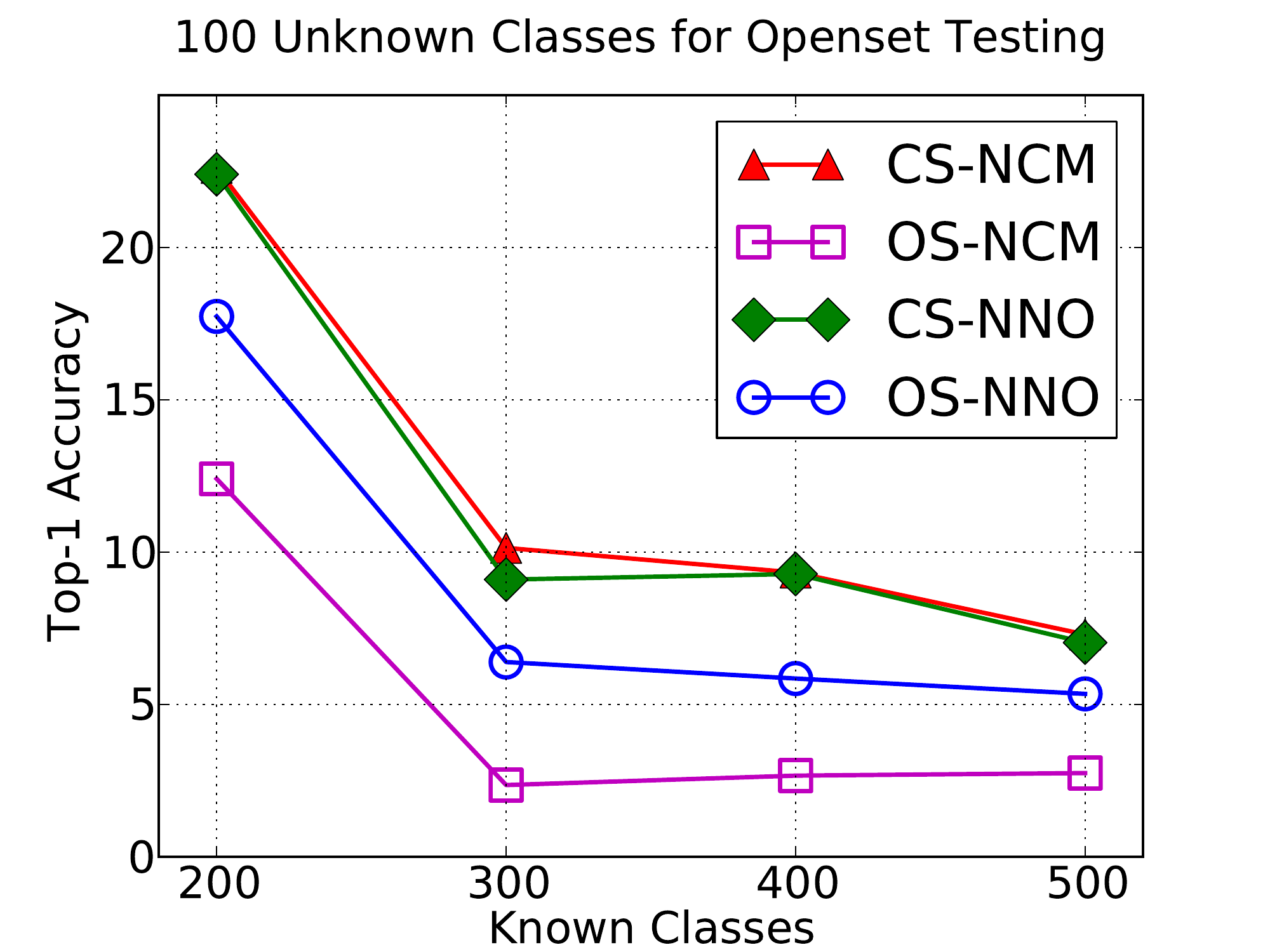}
                \caption{}
                \label{fig:200a}
        \end{subfigure}%
        ~ 
        \begin{subfigure}[b]{0.3\textwidth}
                \includegraphics[width=\textwidth]{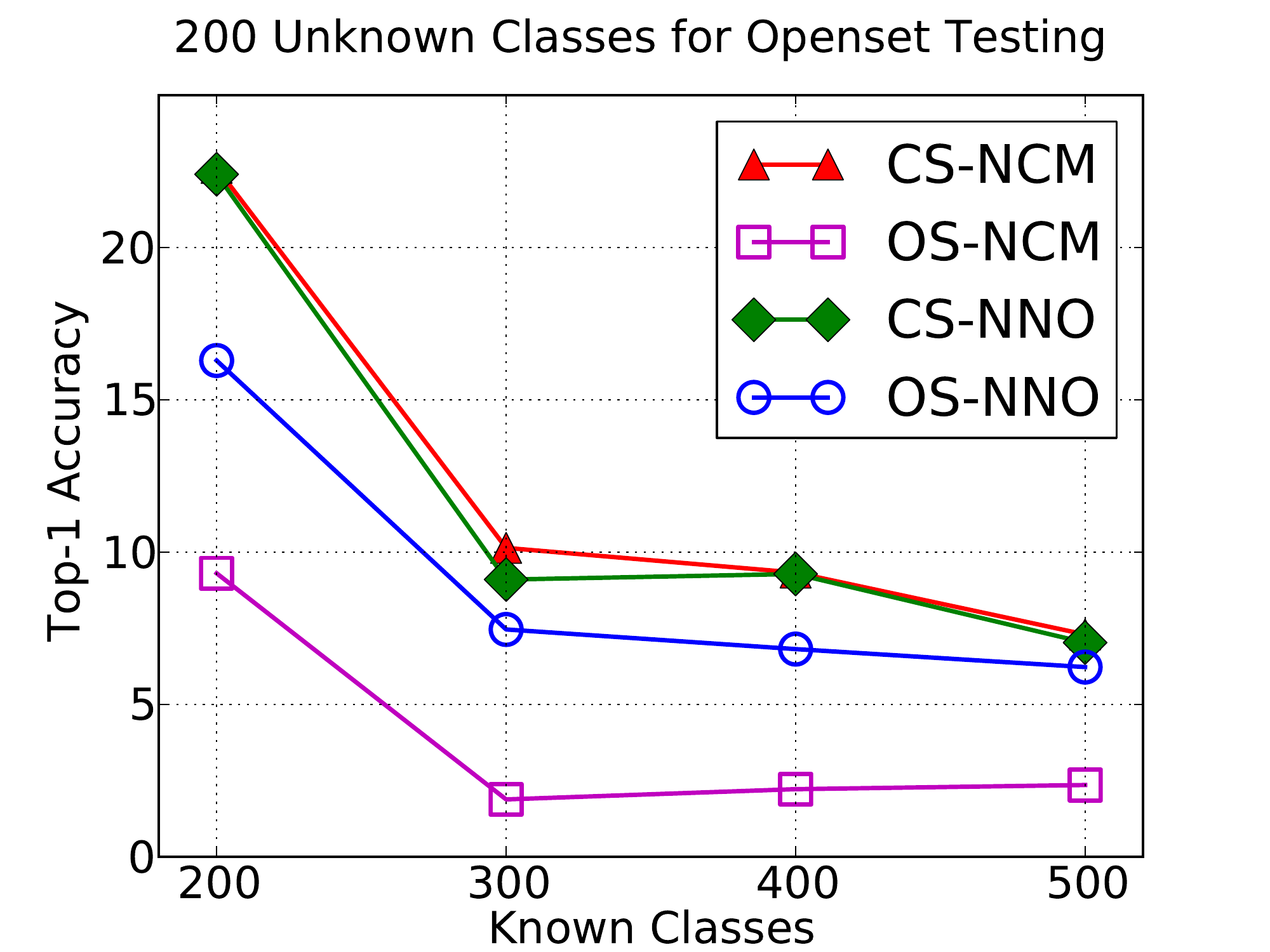}
                \caption{}
                \label{fig:200b}
        \end{subfigure}
        ~ 
        \begin{subfigure}[b]{0.3\textwidth}
                \includegraphics[width=\textwidth]{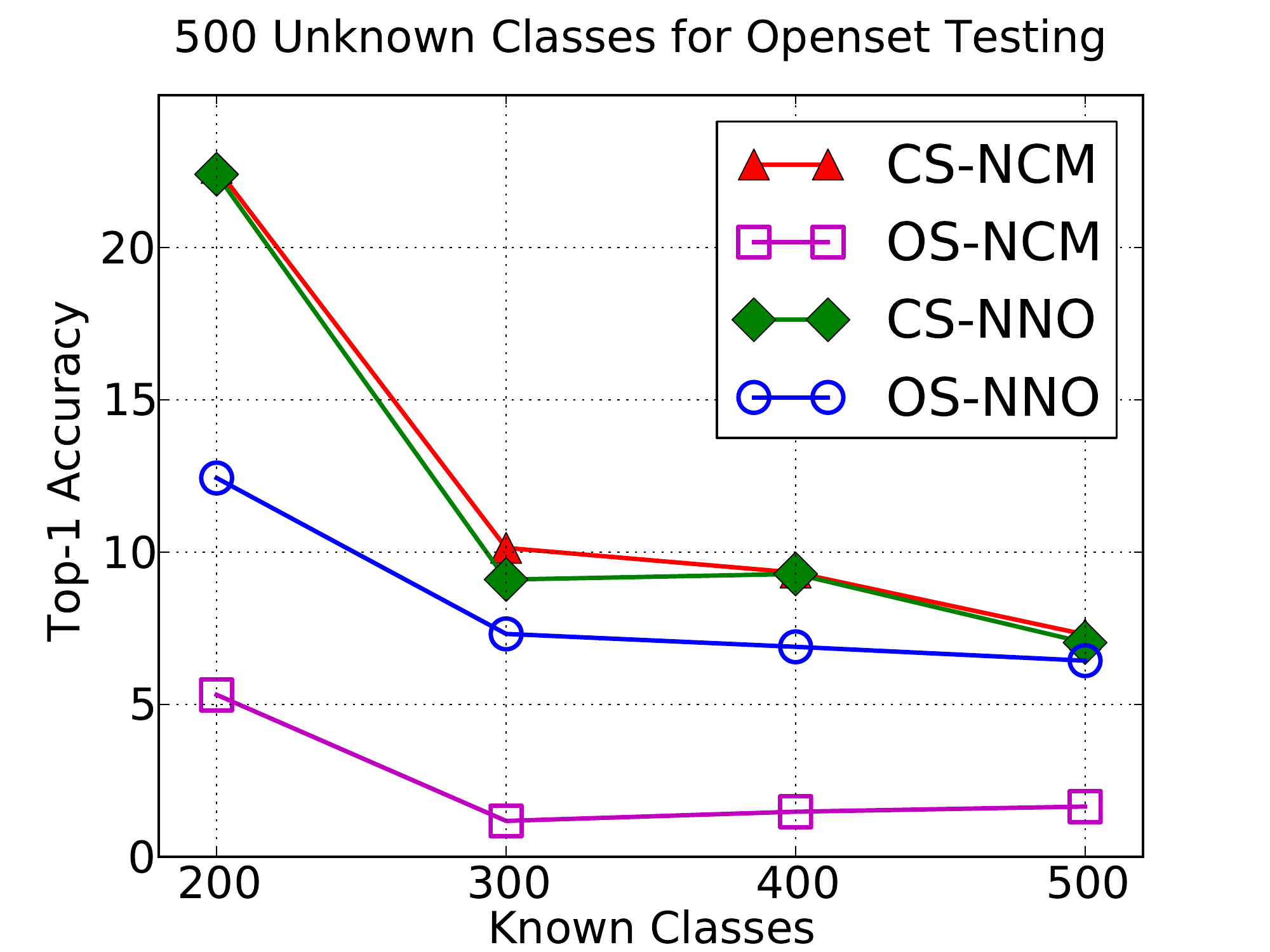}
                \caption{}
                \label{fig:200c}
        \end{subfigure}
        \caption{Open World learning on ILSVRC'10 challenge with 200 initial categories.  With increased size for the Manhanalobis transform via metric learning, the gap between CS-NCM and CS-NNO is decreased as is the gap between closed set and openset. On large scale experiments, OS-NNO continue to outperform OS-NCM. Fig ~\ref{fig:200c} shows metric learning on 200, incremental learning up  to 500 categories and testing with all 1000 categories -- and despite openset testing with 50\% of the classes being unseen, the OS-NNO is performing as well in open set testing as the baseline NCM performs on close set testing (CS-NCM).  
         }
        \label{fig:openworld200}
\end{figure*}

We now compare performance of CS-NCM, OS-NCM and OS-NNO algorithms. In first experiment, we perform metric learning on a relatively few (50) categories to study the validity of the proposed approach. We obtained closed set classification results using CS-NCM which serves as our baseline.  
We add 50 categories incrementally, by updating the system with means of the incoming categories.
To obtain closed set performance, we perform testing with 50, 100, 150 and 200 categories respectively. Table ~\ref{table:protocol} shows the number of training and testing categories used in Fig ~\ref{fig:50a}.
These categories are same as the ones used for training. Open set performance
is obtained by considering an additional 100 unknown categories for testing leading to overall testing with 150, 200, 250 and 300 categories respectively. The results of this experiment are shown in Fig ~\ref{fig:openworld50}. As we add categories to the system in close
 set testing, the performance of both CS-NCM and CS-NNO drop rather gracefully, which is expected. However, in case of open set testing  OS-NCM, the performance drop is drastic because of the unknown categories which NCM was not designed to handel.  More formally, the second
term in Eqn ~\ref{eqn:owerror} dominates the error encountered by the system. 
There is  perforamce drop from CS-NNO to OS-NNO, but nowhere near as dramatic as the NCM drop. 
When the same set of test data is testing with our OS-NNO model, we see significant performance
improvement gains over OS-NCM.  We repeat similar experiment with examples from 200 and 500 unknown categories. We observe OS-NNO consistently performs well on open world recognition task. 

\begin{table}[h]
\begin{tabular}{|l|c|c|c|c|}
\hline
                                                                        & \multicolumn{1}{l|}{Metric Learning}                    & \multicolumn{3}{c|}{Incremental Learning}                                                                                                                                      \\ \hline
\begin{tabular}[c]{@{}l@{}}Training\end{tabular} & 50                                                      & 100                                                      & 150                                                      & 200                                                      \\ \hline
\begin{tabular}[c]{@{}l@{}}Closed Set \\ Testing\end{tabular}           & 50                                                      & 100                                                      & 150                                                      & 200                                                      \\ \hline
\begin{tabular}[c]{@{}l@{}}Open Set \\ Testing\end{tabular}             & \begin{tabular}[c]{@{}c@{}}50 + \\ $100_U$\end{tabular} & \begin{tabular}[c]{@{}c@{}}100 + \\ $100_U$\end{tabular} & \begin{tabular}[c]{@{}c@{}}150 + \\ $100_U$\end{tabular} & \begin{tabular}[c]{@{}c@{}}200 + \\ $100_U$\end{tabular} \\ \hline
\end{tabular}
\caption{The number of classes used for training and testing for the experimental results in Fig ~\ref{fig:50a}. The number of training classes is also the number of known classes. The subscript $\;U$ denotes the number of unknown categories presented to the system during open set testing. This ranges from 100 unknowns for Figs ~\ref{fig:50a}  to 200 unknowns in Fig.~\ref{fig:50b} and 500 unknowns in Fig.~\ref{fig:50c} (.)}
\label{table:protocol}
\end{table}

In second experiment, we consider 200 categories for metric learning and parameter estimtion, and successively add 100 categories in the incremental learning phase. By the end of the learning process, the system needs to learn a total of
500 categories. Open set evaluation of the system is carried out with 100, 200  and 500 unknown categories with results show in Figs  
~\ref{fig:200a}, ~\ref{fig:200b} and ~\ref{fig:200c} respectively.    In final stage  of the learning process i.e 500 categories for training
and 500 (known) + 500 (unknown) categories for open set testing (Fig ~\ref{fig:200c} ), we use all the categories from ImageNet (1000) for our evaluation process. We observe similar rank-ordering of algorithms in this
experiment as in the previous experiment. On the largest scale task involving 500 categories in training and 1000 categories in testing, we observe
almost 74\% improvement of OS-NNO over OS-NCM. We repeated the above experiments over multiple folds and found the standard deviation across folds  to be on the order of $\pm$ 1\% which is not visible in the figure. 

The training time required for the initial metric learning process depends the SGD speed and convergence rate.  We used close to 1M iterations which
resulted in metric-learning time of 15 hours in case of 50 categories and 22
hours in case of metric learning for 200 categories.   Given the metric, the
learning of new classes via the update process is extremely fast as it is simply
computation of means from labeled data.  The majority of time in update process
is dominated by feature extraction and then file I/O, but could easily be in real time. 
Multi-class recognition, including detecting novel classes, is also easily done in real time.

\section{Discussion}

In this work, we formalized the problem of open world recognition, and provide
an open world evaluation protocol.  We extended existing work on NCM classifiers
and showed formally how they can be adapted for open world recognition. The
proposed NNO algorithm consistently outperforms NCM on open world recognition
tasks and is comparable to NCM on closed set -- we gain robustness to the open
world without much sacrifice.  

There are multiple implications of our experiments. First,  we demonstrates suitability
of NNO for large scale recognition tasks in dynamic environments. NNO allows
construction of scalable systems that can be incrementally add classes and that
are robust to unseen categories. Such systems are suitable where minimum down
time is desired.

Second,  as can be seen in Figs. ~\ref{fig:openworld50}, ~\ref{fig:openworld200}
as the number of categories known to the system increases, OS-NNO remains
relatively stable but the closed set performance for CS-NCM and CS-NNO quickly
reduces down to the performance open set. This suggests incrementally adding
more classes in the system is limited by open space risk and that closed set
recognition problem becomes similar to open world recognition problem.  We
conjecture that as the number of classes grow the close world converges to an
open world and thus open world recognition is a natural setting for building
scalable systems.

While we provide one viable approach to extension, the theory herein allows a
broad range of approaches; improved CAP models and better open set probability
calibration should be explored.

Open world evaluation across multiple features for a variety of applications is
an important future work.  Recent advances in deep learning and other areas of
visual recognition have demonstrated significant improvements in absolute
performance.  The best performing systems on such tasks use parallel system and
train for days.  Extending these to incremental open world performance, one may
be able to reuse the deeply learned features with a top layer of open world
multi-class to provide a hybrid solution. While scalable learning in open world
is critical for deploying computer vision applications in the real world, high
performing systems enable adoption by masses. Pushing absolute performance on
large scale visual recognition challenges \cite{berg-imagenet} development of
scalable systems for open world are essentially two sides of the same coin.


\newpage
\clearpage
\section{Supplemental Material : Towards Open World Recognition}

In this supplemental section, we provide additional material to further the reader's understanding of the work on open world recognition, CAP models and the Nearest Non-Outlier algorithm that we present in the main paper.  We present additional experiments on ILSVRC 2010 dataset.  We then present experiments on ILSVRC 2012 dataset to demonstrate that performance gain of OS-NNO over CS-NCM (see fig 3 and 4 in the main paper) are not feature/dataset specific. Finally  first provide algorithmic pseudocode for implementing the NNO algorithm.

\subsection{Experiments on ILSVRC 2010}

	\subsubsection{Thresholding NCM-Softmax for ILSVRC 2010}

\begin{figure}[tbh]
\centering
\includegraphics[width=1\columnwidth]{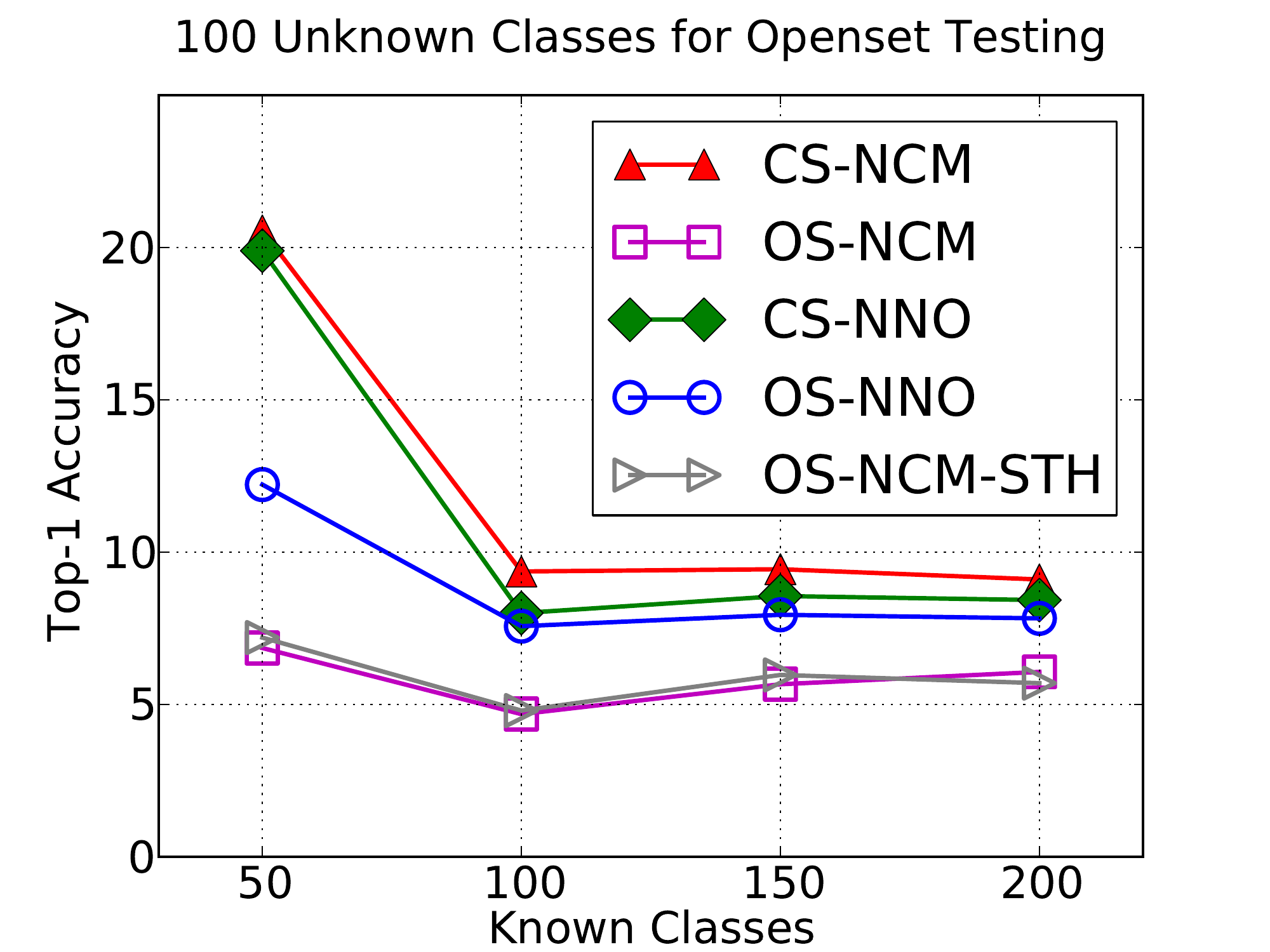}
\caption{Effect of open set performance of thresholding  softmax probabilities. OS-NCM-STH denotes NCM algorithm with open set testing with thresholded softmax probabilities. As can be seen clearly, just thresholding a probability estimates does not produce good open set performance.}
\label{fig:softmax}
\end{figure}

	In section 4.1 of the main paper, we explain the process of rejecting samples from unseen categories to balance open space risk and defined in Eq. 8, a probability function which is thresholded at zero.   At first it might seem like a viable idea to just threshold  the original softmax probability used in NCM.  As explained in the main paper this will fail for open set because the normalization is improper and hence the softmax probability calibration will bias results. To convince the skeptical reader, we add  a small experiment, similar to fig 3a in the main paper, and show the performance of classifying samples as unknown by directly thresholding softmax probabilities. The reader can observe the performance of OS-NCM-STH is similar to OS-NCM and significantly worse than OS-NNO.  Just thresholding the softmax probability is not enough, because its normalization keeps it from decaying as one move away from known data.
	This result confirms the suitability of balancing open set risk with Eq 8,using transformed learned Mahalanobis distance to the NCM. The results from this experiment are shown in fig ~\ref{fig:softmax}. Table 1 lists the different algorithms used.

	\subsubsection{Performance of NNO for different values of $\tau$}
	Section 4.1 and 5.1 in the main paper describes NNO algorithm in detail and steps involved in estimating optimal $\tau$ required to balance open space risk. 
	Section  Alg ~\ref{NNO-pseudocode} illustrates steps involved in developing NNO algorithm for open set. In the experimental results shown in Fig 3 and 4 in the main
	paper, we used optimal $\tau$ for evaluation purpose, which was approximately 5000. In this section, we show the effect of different values of $\tau$ on the performance of OS-NCM to give the reader a feeling for the sensitivity of that parameter.  The optimal value is part of a broad peak, and small changes in $\tau$ have minimal impact. Even changing it by 20\% has only a small impact on open set testing. 
	These results are illustrated with respect to fig 3a in the main paper. In our experiments, we observed similar trends for all other experiments. 
	
\begin{figure}[t]
\centering
\includegraphics[width=1\columnwidth]{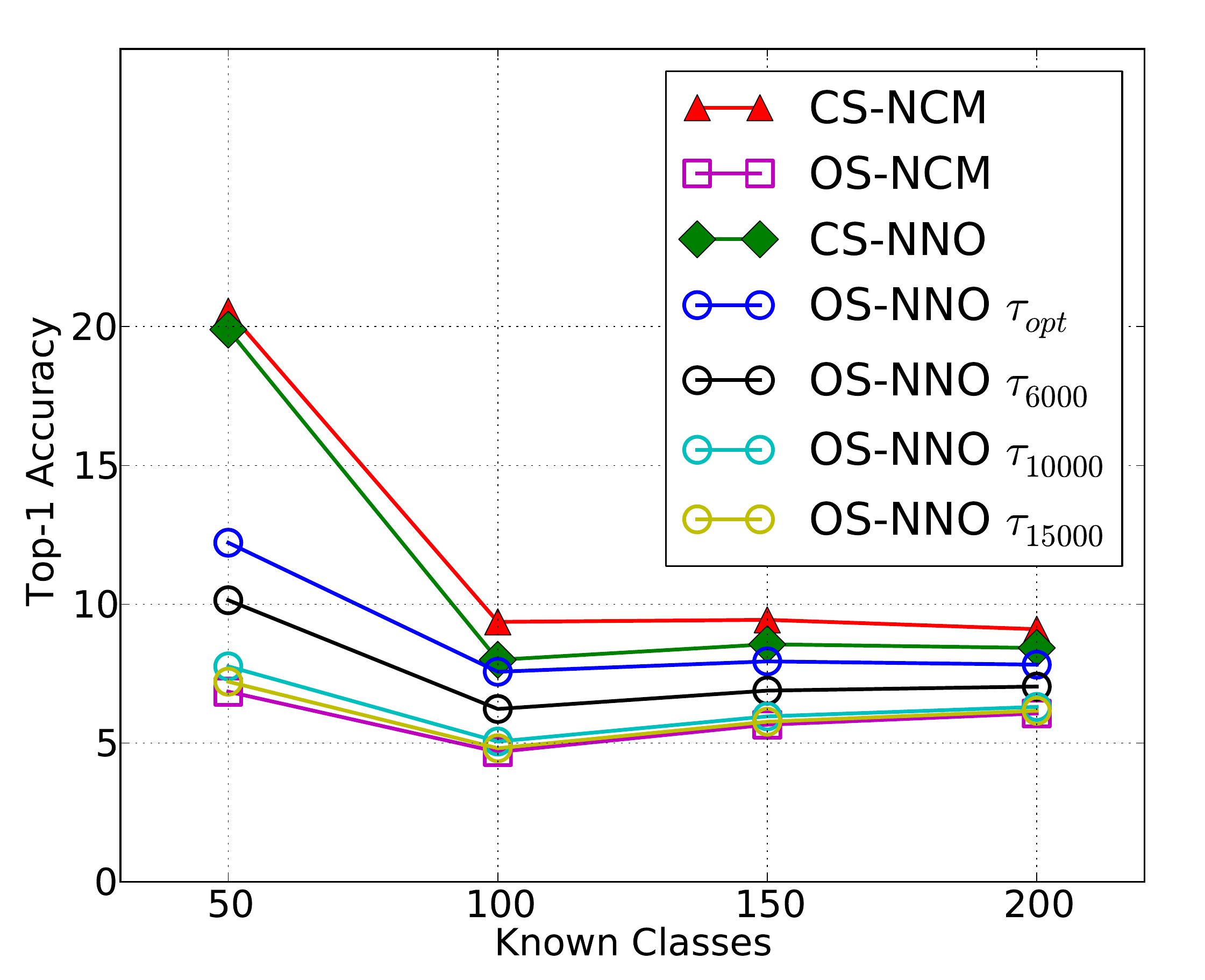}
\caption{The above figure shows the effect of varying threshold $\tau$ on top-1 accuracy on ILSVRC'10 data. The results from CS-NCM, OS-NCM and CS-NNO are same as those shown in fig 3a in the main paper. Here $\tau_{opt} = 5000$, which was the selected threshold for experiments in fig 3a. For a threshold value lower than $\tau_{opt}$, the number of correct predictions retained reduces significantly.}
\label{fig:varythresh}
\end{figure}	
	
	Fig ~\ref{fig:varythresh} shows performance for varying set of $\tau$. $\tau_{opt}$ is the optimal threshold that was selected. We observe that performance of OS-NNO continues to improve as we near
	the optimal threshold. For a threshold value lower than $\tau_{opt}$, the number of correct predictions retained reduces significantly. Thus, a balance between correct predictions retained and unknown categories rejected has to be maintained. This balance is maintained by the selected $\tau_{opt}$. 

\begin{table}[h]
\begin{tabular}{|l|l|}
\hline
Notation & Algorithm                                                                                           \\ \hline
CS-NCM   & \begin{tabular}[c]{@{}l@{}}NCM Algorithm with \\ closed set evaluation\end{tabular}                 \\ \hline
OS-NCM   & \begin{tabular}[c]{@{}l@{}}NCM Algorithm with \\ open set evaluation\end{tabular}                   \\ \hline
CS-NNO   & \begin{tabular}[c]{@{}l@{}}Nearest Non-Outlier Algorithm \\ with closed set evaluation\end{tabular} \\ \hline
OS-NNO   & \begin{tabular}[c]{@{}l@{}}Nearest Non-Outlier Algorithm \\ with open set evaluation\end{tabular}   \\ \hline
OS-NCM-STH   & \begin{tabular}[c]{@{}l@{}} NCM Algorithm Softmax Threshold \\ with open set evaluation\end{tabular}   \\ \hline
\end{tabular}
\caption{The above table shows acronyms used for different algorithms both in the main paper and the supplemental material}
\end{table}

\subsection{Experiments on ILSVRC 2012 Dataset}

\begin{figure*}
        \centering
        \begin{subfigure}[b]{0.3\textwidth}
                \includegraphics[width=\textwidth]{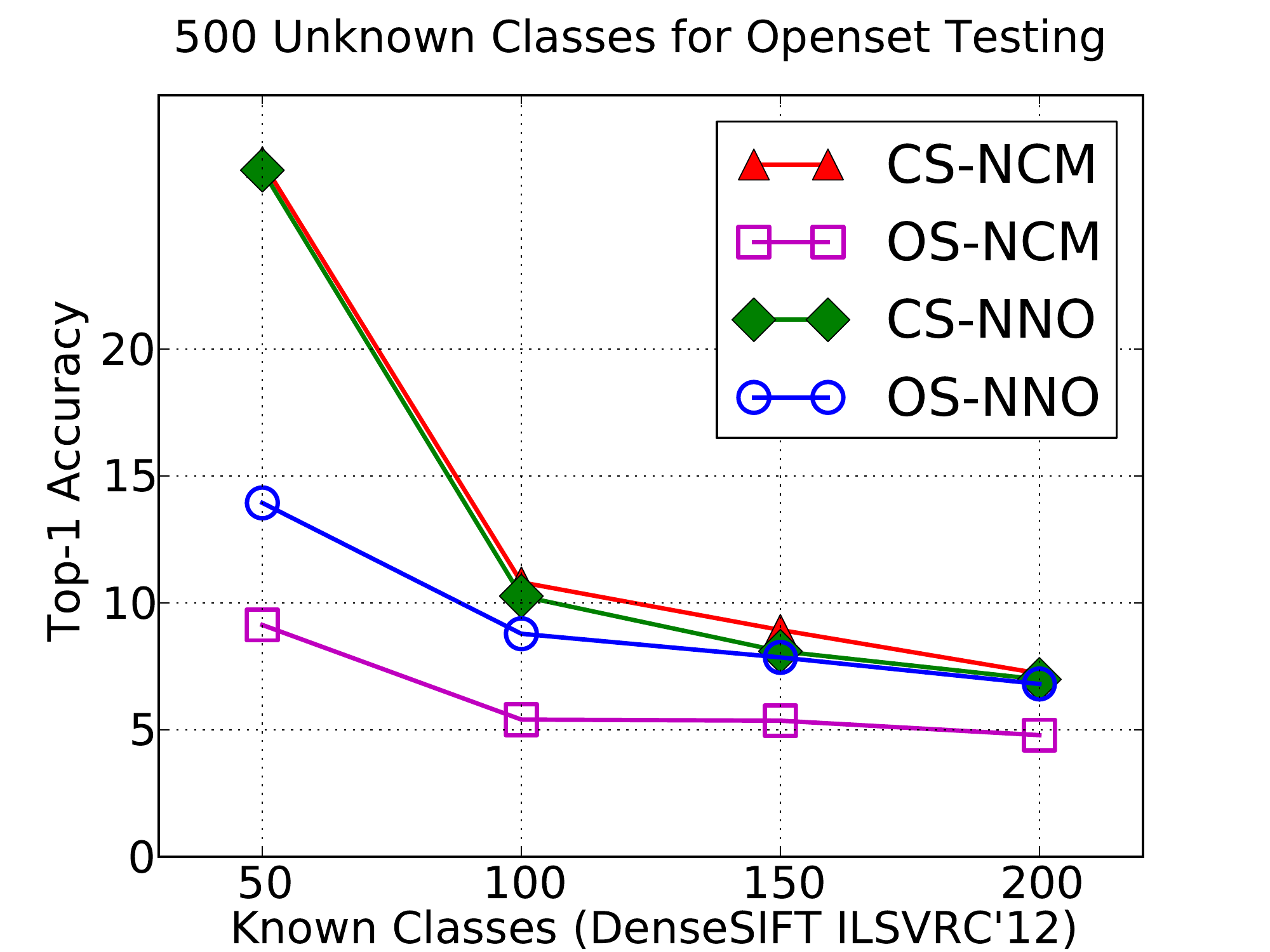}
                \caption{}
                \label{fig:50DENSESIFT12}
        \end{subfigure}%
        ~ 
        \begin{subfigure}[b]{0.3\textwidth}
                \includegraphics[width=\textwidth]{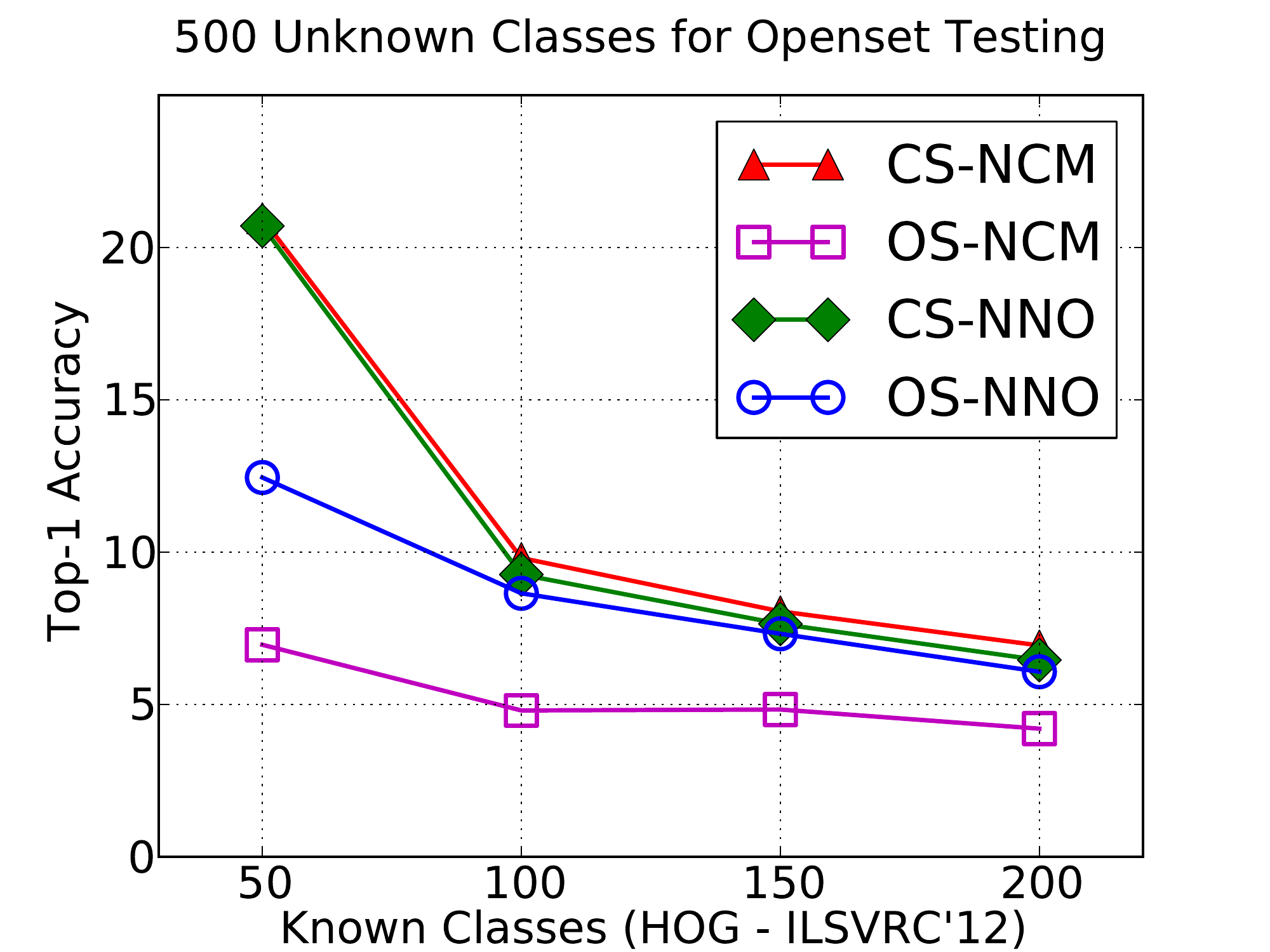}
                \caption{}
                \label{fig:50HOG12}
        \end{subfigure}
        ~ 
        \begin{subfigure}[b]{0.3\textwidth}
                \includegraphics[width=\textwidth]{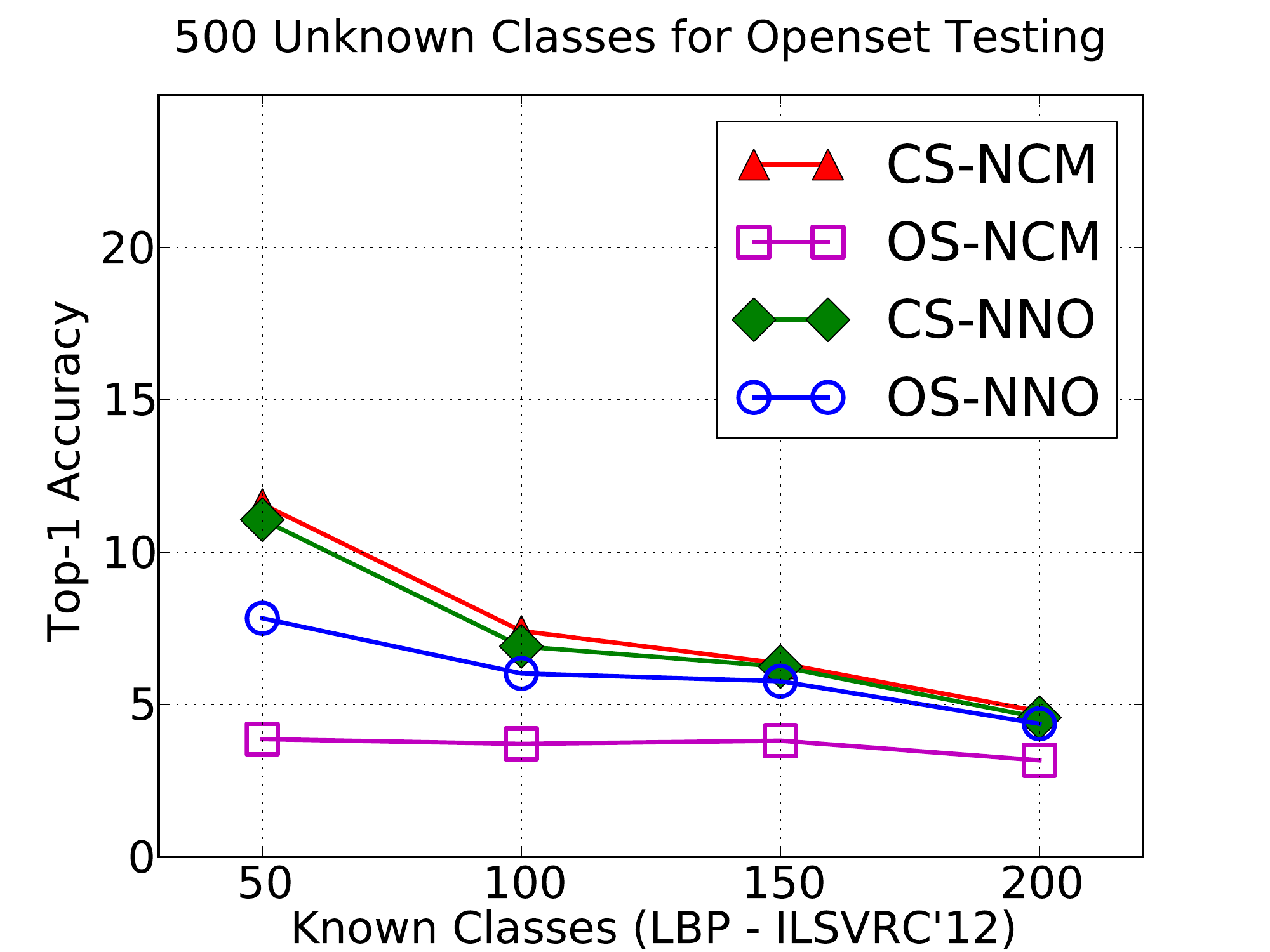}
                \caption{}
                \label{fig:50LBP12}
        \end{subfigure}
        \caption{ The above figure shows experiments on on ILSVRC'12 data. The training data for ImageNet'12 was split into
        train (70\%) and test split (30\%). We show results using three popular features: Dense SIFT ~\ref{fig:50DENSESIFT12}, HOG ~\ref{fig:50HOG12} and LBP ~\ref{fig:50LBP12}. For open set evaluation we use data from 500 unknown categories. This is similar to experiment shown in fig 3c in the main paper. The absolute performance varies from feature to feature, however, we see similar trends in performance as we saw on ILSVRC'10 data.}
        \label{fig:ILSVRC12}
\end{figure*}

As noted in the section 5 (Experiments) in the main paper, we used ILSVRC 2010 dataset because we needed
access to ground truth to for the test set. Ground truth is  was necessary to  perform the open world recognition test protocol, which includes selecting  known and unknown set of categories. In this section, we perform additional experiments 
on ILSVRC 2012 \cite{ILSVRCarxiv14}  \footnote{ILSVRC dataset remained unchanged between 2012, 2013 and 2014} dataset across multiple features to show the effectiveness of NNO algorithm for closed set and
open set tasks does not significantly depend on feature type.

\begin{algorithm*}[t]
\caption{Nearest Non-Outlier Algorithm}
\begin{algorithmic}
\Require $X_k, \mu_k$   \Comment{Initial Training Data $X_k$ from $k$ categories and their means $\mu_k$}

\Function{MetricLearn}{$X_k, \mu_k$}   

\State $W$  = NCMMetricLearn($X_k, \mu_k$)		\Comment{Train NCM Classifier}
\For{$i=1 \to m$}							\Comment{Over multiple folds}	
\State $X_{k_K}, X_{k_U}$ = SplitKnownUnknown($X_k$)	\Comment{Split Training Data into known and unknown set}

\State $\tau_i$ = OpenSetThresh($X_{k_K}, X_{k_U}$)		\Comment{Estimate optimal $\tau_i$ for each split}

\EndFor
\State $\tau$ = $\frac{1}{m}\sum_{i=1}^m \tau_i$			\Comment{Use average $\tau$}
\State NNOModel$_k$ = $[W, \mu_k, \tau]$
\EndFunction

\Require NNOModel$_k$, $X_n, \mu_n$					\Comment{Add additional data $X_n$ from $n$ categories with means $\mu_n$}
\Function{IncrementalLearn}{ NNOModel$_k$, $X_n, \mu_n$}

\State \State NNOModel$_{k+n}$ = $[W, [\mu_k, \mu_n],  \tau]$		\Comment{Update model with means $\mu_n$ }
\EndFunction

\end{algorithmic}
\label{NNO-pseudocode}
\end{algorithm*}

Since ground truth is not available for ILSVRC'12 dataset, we split the training data provided by
the authors into training and test split.  The number of categories is the same, this just limits the number of images per class used.  We use 70\% of training data to train models and 30\% of the data for evaluation. This process is repeated over multiple folds. Once the data is split into training and test split, remaining
procedure for metric learning and incremental learning is followed similar to
that in section 5 (Experiments) in the main section. We conduct similar 2 sets of experiments on ILSVRC'12 data:
metric learning with 50 and 200 initial categories as shown in Figs 3 and 4 in the main paper. The closed set and open open set testing is conducted in similar manner as well. While the open world experimental setup for ILSVRC'12 is not ideal because of the smaller number of images per class, the goal of this experiment is to show that the advantages of NNO are not feature dependent.

We use pre-computed features as provided on cloudcv.org \cite{CloudCV}. We consider three set of features as follows:

\begin{enumerate}
 \item{ \textbf{Dense SIFT:}} SIFT descriptors are densely extracted \cite{Lazebnik-sift} using a flat window at two scales (4 and 8 pixel radii) on a regular grid at steps of 5 pixels. The three descriptors are stacked together for each HSV color channels, and quantized into 300 visual words by k-means. The features used in the main paper are similar to these features, except that in the main paper, dense SIFT features were quantized into 1000 visual words by k-means. 
\item \textbf{Histogram of Oriented Gradients (HOG):} HOG features are used in wide range of visual recognition tasks \cite{hog-dalal}. HOG features are densely extracted on a regular grid at steps of 8 pixels. HOG features are computed using code provided by \cite{Felzenszwalb-hog}. This gives a 
 31-dimension descriptor for each node of the grid. Finally, the features are quantized into 300 visual words by k-means.
 \item \textbf{Local Binary Patterns (LBP):} Local Binary Patterns (LBP) \cite{lbp} is a texture feature based on occurrence histogram of local binary patterns. It has been widely used for face recognition
 and object recognition. The feature dimensionality used was 59. 
 \end{enumerate}

Results using Dense SIFT, HOG and LBP features are shown in figures ~\ref{fig:50DENSESIFT12}, ~\ref{fig:50HOG12} and ~\ref{fig:50LBP12} respectively.
The absolute performance with Dense SIFT features is the best, followed by HOG and LBP.  The Dense SIFT is very similarly to the results on ILSVRC 2010.  Moreover, from these
experiments we observe similar trends across all features to the trends seen in Figs 3 and 4 in the main paper. We see that
as closed set performance of CS-NCM and CS-NNO is comparable while OS-NCM suffers significantly when
tested with unknown set of categories. We continue to see significant gains of OS-NNO over OS-NCM across
HOG and dense SIFT features. We also observe the trend where as we add more categories in the system,
the closed set and open set performance begin to converge. Thus, it is reasonable to conclude that the performance
gain seen in terms of OS-NNO is not feature dependent. These observations are consistent with our observations
from experiments on ILSVRC'10 data.

\subsection{Algorithmic Pseudocode for Nearest Non-Outlier (NNO)}

In this section, we provide pseudocode for Nearest Non-Outlier algorithm as described in section 4.1 in the main
paper. The algorithm proceeds in multiple steps. In the first step, features are normalized by the mean
and standard deviation over the starting subset. The initial set of features is used to perform metric learning.
Following this step, threshold $\tau$ for open set NNO is estimated  using per class decisions using per Eq. 8 in the main paper and a cross class validation procedure of \cite{boult-pisvm}  training data splits. The complete pseudocode is given in Alg ~\ref{NNO-pseudocode}

\subsection{Acknowledgement}
We would like to thank Thomas Mensink (ISLA, Informatics Institute, University of Amsterdam) for sharing code and Marko Ristin (ETH Zurich) for sharing features. The work was carried out with the help of NSF Research Grant IIS-1320956 (Open Vision - Tools for Open Set Computer Vision and Learning) and UCCS Graduate School Fellowship. 

{\small
\bibliographystyle{ieee}
\bibliography{inclabeltree}
}

\end{document}